\documentclass[10pt,letterpaper]{article}

\usepackage{cogsci}

\cogscifinalcopy 
\usepackage{microtype}
\usepackage{times}
\usepackage{subfig}
\usepackage{hyperref}
\usepackage{amsmath}
\usepackage{wrapfig}
\usepackage{amssymb}
\usepackage{xcolor}
\usepackage{booktabs}
\usepackage{natbib}
\usepackage{latexsym}
\usepackage{graphicx}
\usepackage{booktabs}
\usepackage{multicol}
\usepackage{multirow}

\usepackage{pslatex}
\usepackage{float} 

\newcommand*\samethanks[1][\value{footnote}]{\footnotemark[#1]}

\title{Structural Inductive Biases in Emergent Communication}
 
\author{
  Agnieszka S\l{}owik\thanks{Equal Contribution.} \\
  Department of Computer Science and Technology,\\
      University of Cambridge \\
  \texttt{agnieszka.slowik@cl.cam.ac.uk} \\ \And
  \textbf{Abhinav Gupta}\samethanks  \\
  MILA \\
  \texttt{abhinavg@nyu.edu}
  \\\AND
  \textbf{William L. Hamilton} \\
  School of Computer Science, \\
  McGill University
\\\And
\textbf{Mateja Jamnik} \\
Department of Computer Science and Technology,\\
University of Cambridge
    \\\AND
\textbf{Sean B. Holden} \\
Department of Computer Science and Technology,\\
     University of Cambridge
    \\\And
\textbf{Christopher Pal} \\
Polytechnique Montr\'eal \\
ServiceNow 
}

\begin{document}

\maketitle

\begin{abstract}
In  order  to  communicate, humans flatten a complex representation  of ideas and their attributes into a single word or a sentence. We investigate the impact of representation learning in artificial agents by developing graph referential games. We empirically show that agents parametrized by graph neural networks develop a more compositional language compared to bag-of-words and sequence models, which allows them to systematically generalize to new combinations of familiar features. 

\end{abstract}

\section{Introduction}

Human language is characterized by the ability to generate a potentially infinite number of sentences from a finite set of words. In principle, humans can use this ability to express and understand complex hierarchical and relational concepts, such as kinship relations and logical deduction chains. Existing multi-agent communication systems fail to compositionally generalize even on constrained symbolic data \citep{kottur2017natural}.
For instance, if a person knows the meaning of utterances such as `red circle' and `blue square', she can easily understand the utterance `red square' even if she has not encountered this particular combination of shape and color in the past. This type of generalization capacity is referred to as \textit{compositionality} \citep{Smith:2003:ILF:963725.963729,andreas2018measuring,baroni_linguistic_2019} or \textit{systematic generalization} \citep{bahdanau2018systematic}.

The ability to represent complex concepts and the relationships between them in the manner of a mental graph was found to be one of the key factors behind knowledge generalization and prolonged learning in humans \citep{Bellmundeaat6766}. Through communication, humans flatten the non-Euclidean representation of ideas into a sequence of words. Advances in graph representation learning \citep{kipf2017semi} and sequence decoding \citep{sutskever_sequence_2014,cho-etal-2014-learning} provide the means for simulating this \textit{graph linearization} process in the artificial agents.

In \emph{emergent communication} studies, learning agents develop a communication protocol from scratch through solving a shared task, most commonly a \emph{referential game} \citep{DBLP:conf/iclr/LazaridouHTC18, guo2020inductive}. Modelling communication as an interactive, goal-driven problem mitigates some of the issues observed in supervised training, such as sample inefficiency and memorizing superficial signals rather than discovering the real factors of variation in the data \citep{lake_ullman_tenenbaum_gershman_2017,baroni_linguistic_2019}.

In this work, we analyze the effect of varying the input representation and the corresponding representation learning method on compositional generalization in referential games. Given the recent resurgence of strong structural bias in neural networks, we evaluate the hypothesis that graph representations encourage compositional generalization \citep{DBLP:journals/corr/abs-1806-01261} in emergent communication. 

\section{Related work}
In prior work using referential games, input data is represented as sequences and bags-of-words \citep{DBLP:conf/iclr/LazaridouHTC18,bouchacourt_how_2018} with few instances of using images as input \citep{DBLP:conf/iclr/LazaridouHTC18,DBLP:conf/iclr/EvtimovaDKC18} and one instance of using graph encoders without a sequence decoder \citep{slowik2020towards,gupta2020analyzing}. Notably, more information in the input does not necessarily lead to a more compositional language or higher generalization as shown in \citep{DBLP:conf/iclr/LazaridouHTC18}. \citet{DBLP:journals/corr/abs-1806-01261} claim that graph representations induce stronger compositional generalization across a variety of reasoning and classification tasks. Recent work in relational reasoning \citep{sinha-etal-2019-clutrr} supports this claim.

\section{Environment}
\subsection{Referential games}
A referential game involves two agents of fixed roles: the \textit{speaker} and the \textit{listener}. The speaker has access to the input data (\emph{target}), which the listener cannot directly observe. The speaker sends a message describing the target and the listener learns to recognize the target based on the message. In the referential games studied in this work, the listener receives the message and a set consisting of the target and \textit{distractors}, new objects sampled without replacement from the target data distribution. The agents are rewarded if and only if the receiver recognizes the target, and not for the messages sent. In the existing work, target and distractors are represented as bags-of-words, sequences or images. See the \textbf{Training} section for details on training.

\subsection{Graph referential games}

\begin{figure}
    \centering
    \includegraphics[width=\linewidth]{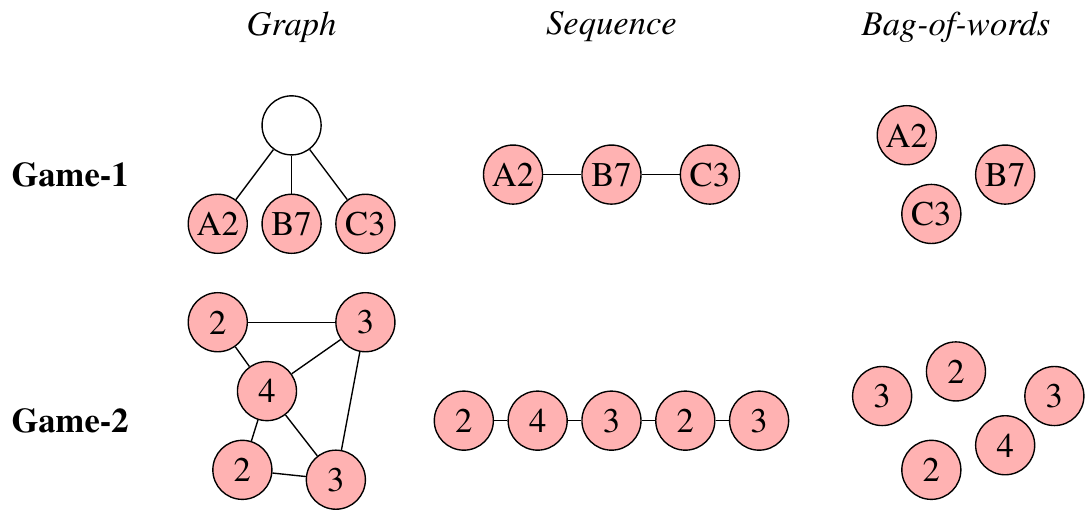}
    \caption{Structural biases in the input. Baselines of sequences and bags-of-words are constructed similarly as in the existing work on emergent communication \citep{kottur2017natural,DBLP:conf/iclr/LazaridouHTC18}. }
    \label{fig:input}
\end{figure}

\begin{figure*}[t!]
    \centering
    \subfloat[9 distractors]{{\includegraphics[scale=0.85]{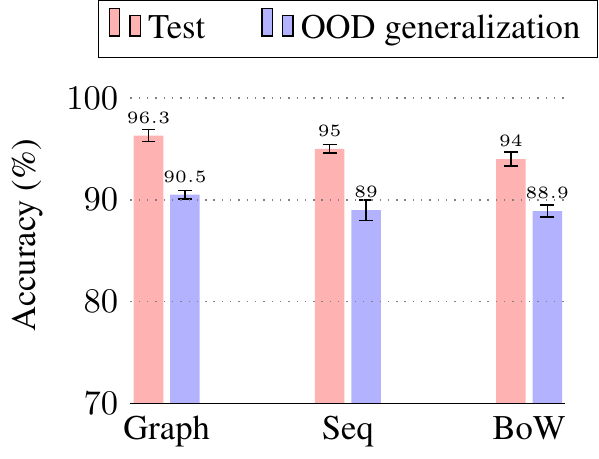}}}
     \hfill
     \subfloat[19 distractors]{{\includegraphics[scale=0.85]{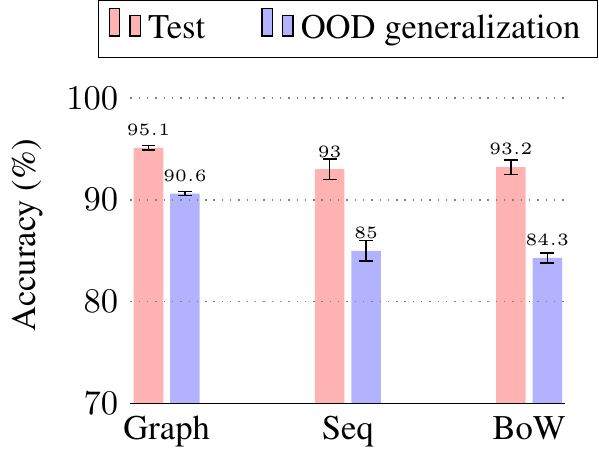}}}
     \hfill
     \subfloat[49 distractors]{{\includegraphics[scale=0.85]{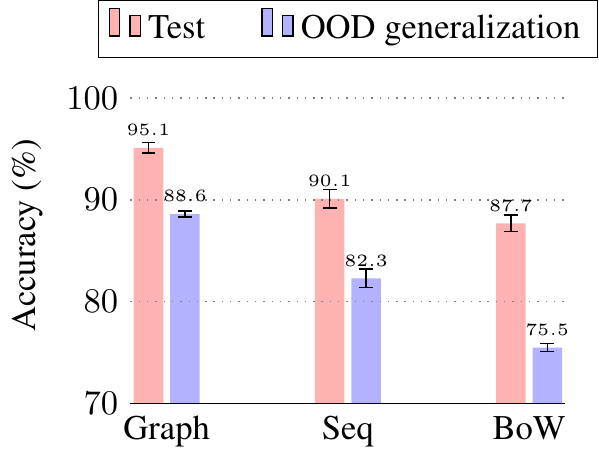}}}
    \caption{Standard test accuracy (Test) and out-of-domain (OOD) generalization in Game-2. Number of nodes, message length and vocabulary size are equal to 25. We report mean and standard deviation of three runs.
    }
     \label{fig:ood-comparison}
\end{figure*}

We extend existing referential games by developing graph referential games: \emph{Game-1} and \emph{Game-2}. We include corresponding representations of a lower degree of structure in order to study the effect of varying the input representation (Figure \ref{fig:input}). 

 \paragraph{Game-1: hierarchy of concepts and properties}
In this game, we construct a tree from a vector of $[p_{1}, p_{2}, \dots, p_{n}]$ (which we will refer to as \emph{perceptual dimensions}) where $n$ corresponds to the number of properties and $p_1, p_2, \dots, p_n$ denote the number of possible types per property. Each tree has the same number of properties $n$ and they only differ in the property values. Formally, each tree is an undirected graph $\mathcal{G}(\mathcal{V},\mathcal{E})$ where $\mathcal{V}$ is the set of all nodes representing unique properties along with a `central' node, and $\mathcal{E}$ is the set of edges. The central node corresponds to a conceptually more abstract representation of the input, which is initially empty and then learned by a graph agent based on the object properties. The node features consist of a concatenation of the property encoding and the type encoding (represented as one-hot vectors).

\paragraph{Game-2: relational concepts} 
In this game, we evaluate the graph agents using arbitrary undirected graphs of varying edges. Such graphs can be used to represent relations between arbitrary entities, e.g.~connections between users of a social media platform. In Game-2, each undirected graph $\mathcal{G}(\mathcal{V},\mathcal{E})$ is defined over the set of nodes $\mathcal{V}$ and the set of edges $\mathcal{E}$. In a given instance of the game, $|\mathcal{V}|$ is fixed for all targets and distractors. The number of edges varies across the graphs. We add a self-loop to each node to include its own features in the node representation aggregated through message passing. We use node degrees converted to one-hot vectors as the initial node features.

\paragraph{Baseline agents}
In each instance of the game, the speaker is parametrized by an encoder-decoder architecture and the listener is implemented as a classifier over the set consisting of the target and distractors.
We use generic Bag-of-Words2Sequence and Sequence2Sequence (Seq2Seq) \citep{sutskever_sequence_2014} models over the node features as baseline speakers, and corresponding classifiers as listeners.

\paragraph{Graph agents}

In order to handle graph input, the speaker and the listener are parametrized using a graph encoder. The speaker additionally uses a sequence decoder to generate a message. The graph encoder first generates node embeddings for each node, and then it uses them to construct an embedding of the entire graph. The sequence decoder takes the graph embedding as input and generates a message. A graph encoder consists of a node representation learning method and a graph pooling method. Node representations are computed for each node $v_{i}$ through neighborhood aggregation that follows the general formula
$
    h_{v_i}^{(l+1)} = \mathsf{ReLU}\left(\sum_{j \in N_i} h_{v_j}^{(l)}W^{(l)}\right),
$
where $l$ corresponds to the layer index, $h_{v_i}$ are the features of the node ${v_i}$, $W$ refers to the weight matrix, and $N_{i}$ denotes the neighborhood of the node $v_i$. We compare a Graph Convolutional Network (GCN) \citep{kipf2017semi}) with GraphSAGE \citep{hamilton_inductive_2017}, an extension of GCN which allows modifying the trainable aggregation function beyond a simple convolution. A graph embedding is obtained through a linear transformation of the node features. A graph embedding vector in our graph-to-sequence implementation of the speaker corresponds to the context vector in the Seq2Seq implementation. Similarly as in Seq2Seq architectures, the sequence decoder in graph-to-sequence outputs a probability distribution over the whole vocabulary for a fixed message length which is then discretized to produce the message.

\section{Training}
\label{training}
The speaker produces a softmax distribution over the vocabulary $V$, where $V$ refers to the finite set of all distinct words that can be used in the sequence generated by the speaker. Similar to \citep{sukhbaatar2016learning,mordatch_emergence_2017}, we use the `straight through' version of Gumbel-Softmax \citep{jang_categorical_2016,maddison2016concrete} during training to make the message discrete and propagate the gradients through the non-differentiable communication channel. At test time, we take the \texttt{argmax} over the whole vocabulary.

In our graph referential games, the listener receives the discretized message $m$ sent by the speaker along with the set of distractors $K$ and the target graph $d^*$. The listener then outputs a softmax distribution over the $|K|+1$ embeddings representing each graph. The speaker $f_\theta$ and the listener $g_\phi$ are parametrized using graph neural networks. We formally define it as follows:
$$
   m(d^*) = \texttt{Gumbel-Softmax}(f_\theta(d^*)) 
$$
$$
    o(m, \{K, d^*\}) = g_\phi(m, \{K, d^*\})\\
$$

We used Deep Graph Library \citep{wang2019dgl}, EGG \citep{kharitonov-etal-2019-egg} and PyTorch \citep{pytorch_NIPS2019} to build graph referential games. We generate 40000 train samples, 5000 validation samples and 5000 test samples in each game.

\section{Experiments \& Analysis}
We investigate three questions:
\begin{itemize}
    \item What is the effect of data representation and the corresponding representation learning models on the compositionality of the emerged language (\textbf{Topographic similarity})?
    \item What is the effect of structural biases on the ability to generalize to previously unseen combinations of familiar features (\textbf{Out-of-domain generalization})?
    \item Can the listener identify the target if it receives a distorted message? (\textbf{Do agents rely on the communication channel in solving the game?}).
\end{itemize}
\subsection{Qualitative analysis}
\begin{table}[ht]
\centering
{
\begin{tabular}{cccc}
\multicolumn{1}{c}{\multirow{2}{*}{\bfseries Sample input data}} & \multicolumn{3}{c}{\bfseries Data representation} \\ \cmidrule(r){2-4}
& \emph{Bag-of-words} & \emph{Sequence} & \emph{Graph} \\
\toprule
A2 B4 C6 & [1 4 4] & [5 1 3] & [7 1 4] \\
A2 B4 C5 & [1 0 0] & [8 3 2] & [4 7 2] \\
A2 B2 C6 & [6 4 1] & [9 9 1] & [9 4 1] \\
A5 B4 C6 & [8 8 9] & [3 5 2] & [6 6 1] \\
\end{tabular}
}
\caption{Qualitative samples of messages.}
\label{table:qualitative_samples}
\end{table}
In Table~\ref{table:qualitative_samples}, we show sample messages generated in Game-1. Similarly as in Figure~\ref{fig:input}, we represent input properties using capital letters and property types using numbers. We see that in the messages generated by a Graph2Seq model varying one input property (e.g. replacing C6 with C5) changes only one symbol in the message (the speaker replaces the word 1 with 2 in the utterance). In Seq2Seq, changing one symbol leads to a change of two symbols on average in the transmitted messages. In the example, the vocabulary size is $10$ with message of fixed length $3$ and the perceptual dimensions being $[10, 6, 8]$. When using larger input in Game-1, we experimented with the message length-to-vocabulary size trade-off (details in the \textbf{Hyperparameters} section) and we found that longer messages are less compositional. In Game-2 (Figure 2), the speaker learns to describe 25 properties in 25 words. We found that for messages of the length 25, sample messages are difficult to interpret qualitatively, as all representations lead to languages that are order invariant with respect to words (see Table~\ref{table:qualitative_samples}).

\subsection{Topographic similarity}
\label{toposim}

We use \emph{topographic similarity} as a measure of language compositionality, following a common practice in the domain of referential games \citep{DBLP:conf/iclr/LazaridouHTC18,li_ease--teaching_2019}. We compute a negative Spearman correlation between all the possible pairs of target objects and the corresponding pairs of emerged messages. We use cosine similarity in the input space and Levenshtein distance in the message space. For graph representations we concatenate the node features in the same order as in sequences and bags-of-words for a fair comparison.

 \begin{figure}[ht]
     \centering
     \includegraphics[width=\linewidth]{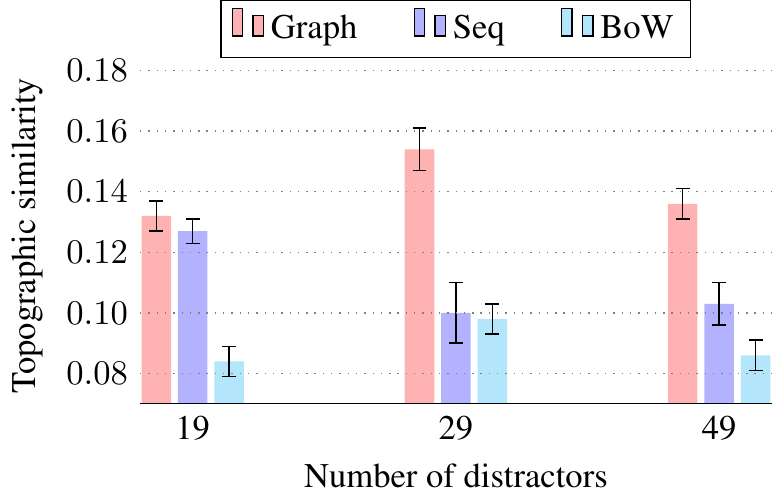}
         \caption{Topographic similarity in Game-1 with perceptual dimensions [10, 6, 8, 9, 10], a message length of size 3, and a vocabulary size of 50. We report mean and standard deviation across five random seeds.
         }
     \label{fig:topo}
 \end{figure}
 
\begin{figure*}[ht]
    \centering
    \includegraphics[width=\textwidth]{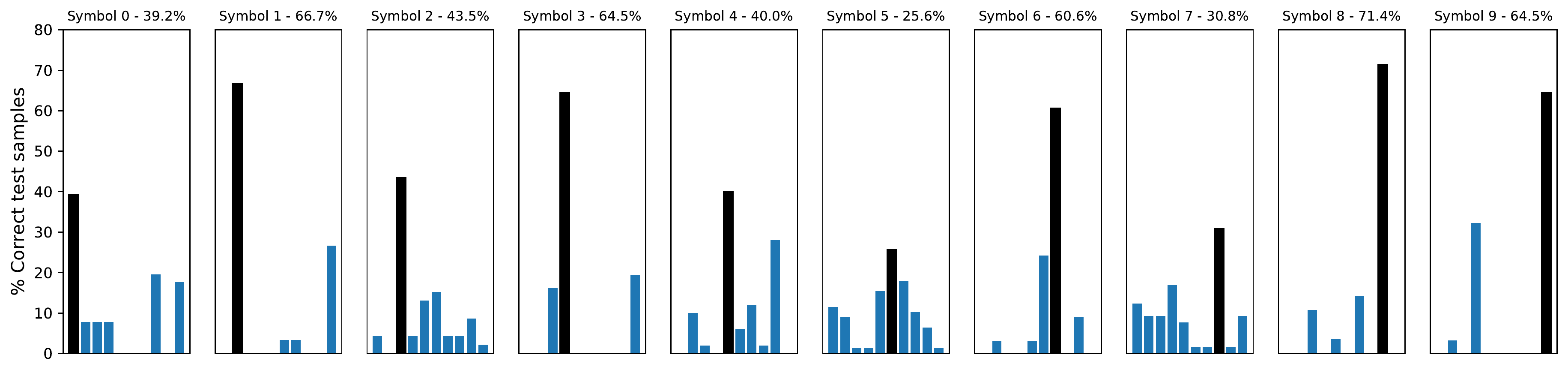}
    \caption{Robustness of the communication protocol. We use trained agents and emerged messages from Game-1 with perceptual dimensions [10, 6, 8], a message length of size 3, and a vocabulary size of 10.}
    \label{fig:robust-graphs}
\end{figure*} 

Figure~\ref{fig:topo} shows topographic similarity (TS) of all three data representations in Game-1. All representations lead to positive values of TS, which implies that there is a positive correlation between the input features and the messages emerging in the game. On average, graph representations lead to a more compositional language (measured by greater TS) than bag-of-words and sequence representations. This effect is observed for instances of Game-1 with different numbers of distractors (19, 29, 49). We hypothesize that in more difficult communication scenarios (e.g., if the number of distractors increases), structured representations are needed for emergence of a compositional language. 
 
\subsection{Out-of-domain generalization}
\label{ood}

In Game-2, we compare sequences, bag-of-words and graph representations in terms of test accuracy (using a 60\% / 20\% / 20\% train/valid/test split) and \emph{compositional generalization} (OOD generalization). In evaluating OOD generalization, the agent sees an unseen graph/sequence/bag-of-words with a new combination of seen features. Intuitively, the agents only learn a ``red dotted circle'', ``blue dotted square'',``yellow dashed star'' during training, and at test time we check if they can correctly identify the target: ``red dashed square''. Since both the target and distractors are new to the agents, both agents are tested for their ability to disentangle input properties. 

We investigated test accuracy and OOD generalization for an increasing number of distractors (9, 19, 49). Accuracy of a random guess in these games is, respectively, 10\%, 5\% and 2\%. Consequently, the game becomes significantly more complex as the number of distractors increases. Figure \ref{fig:ood-comparison} shows the effect of structural priors on test accuracy and compositional generalization. Graph representations consistently lead to a better generalization to new combinations of familiar features, and this effect increases with the complexity of the game. Since for each representation we use generic agents of a comparable expression power, we hypothesize that graph representations lead to a more compositional language, especially in more difficult communication scenarios.

\subsection{Do agents rely on the communication channel in solving the game?}
\label{robustness}

We analyzed whether the communication channel is crucial in learning to recognize the target among distractors. Figure \ref{fig:robust-graphs} shows the results of this analysis over the entire set of targets and messages developed by graph learning agents. We show the results for messages in the form of $\{m_1, m_2, m_3\}$ (message length $\mathit{ml}=3$, $m_i$ corresponds to individual symbols in the message, $i=1,2,3$) and vocabulary size $\mathit{vs}=10$. In each subplot, the title corresponds to the symbol $m_1$ in the original message developed through playing the game. The dark bar corresponds to the number of correctly classified samples given the original message (also included in the subplot title, e.g.~Symbol 0: 39.2\%). We generate the remaining results by replacing the first symbol $m_1$ in the messages with each of the remaining symbols from the vocabulary. The rest of the message (symbols $m_2$ and $m_3$) remains fixed.

We observe that in all emerged messages distorting $1/3$ of the symbols leads to a decrease in test accuracy. In all cases, the highest test accuracy is dependent on the target encoding produced by the graph speaker through playing the game.

\section{Ablation studies on Graph Neural Networks}
\label{ablation}
GraphSAGE learns aggregator functions that can induce the embedding of a new node given its features and the neigborhood, without re-training on the entire graph. The GraphSAGE encoders are thus able to learn dynamic graphs. In this paper, we experiment with commonly used `mean', `pool' and `gcn' aggregator types.

\textit{Pooling methods:} In order to compute the graph embedding, we experimented with the standard graph pooling methods: mean, sum and max functions. We found that the sum pooling gave a significant boost in performance, and thus we use sum pooling throughout the experiments presented in this paper. 

\textit{Encoder networks:} We also experimented with two popular graph neural networks to compute the graph encoding, namely GraphConv and SAGEConv. We did not find a significant difference in performance between the two models. 

\textit{Aggregator types:} Another axis of variation is the aggregator type used in GraphSAGE and we found that the effect of all types- `mean', `pool' and `gcn' is the same across both games.

\begin{figure}
    \centering
    \includegraphics[width=\linewidth, height=5cm]{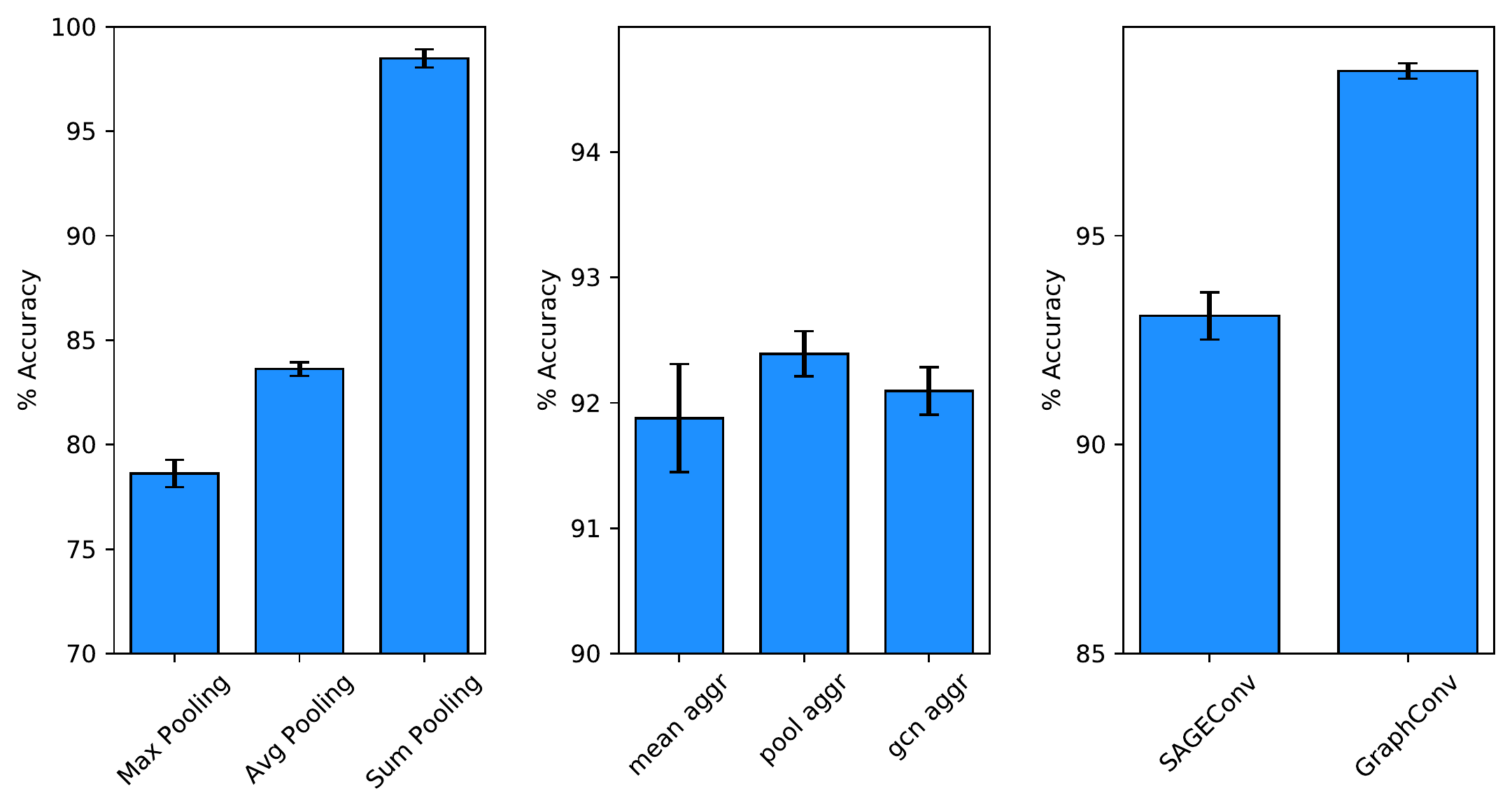}
    \caption{Ablation studies on different parameters of Graph Neural Network. The y-axis represent the test accuracy. All the runs are averaged across three different random seeds and standard error bars are shown.}
    \label{fig:graph-ablation}
\end{figure}

\subsection{Hyperparameters}
\label{hyperparams}
In this work, we studied generic architectures and the main axis of variation was the degree of structure in the input representation and the corresponding representation learning method. In the experiments reported in this paper, we used the a range of hyperparameters listed below (bold indicates the ones used for the charts above) that were manually tuned using accuracy as the selection criterion:

\begin{itemize}
\item Number of distractors: 1, 2, 4, \textbf{9, 19, 29, 49}
\item Vocab Size of the message: \textbf{10, 25, 50}, 100
\item Max length of the message: \textbf{3}, 4, 5, \textbf{10, 25}
\item Number of layers in Graph NN: 1, \textbf{2}, 3
\item Size of the hidden layer: 100, \textbf{200}
\item Size of the message embedding: \textbf{50}, 100
\item Learning Rate: 0.01, \textbf{0.001}
\item Gumbel-Softmax temperature: \textbf{1.0}
\end{itemize}
For more details, please refer to the codebase. We used Nvidia V100 and Titan RTX GPUs for running our experiments. On average, the models were trained for 3-5 days depending on the $\#$distractors and the type of data representation used with Graph2Sequence models taking the longest and Bag2Sequence models the shortest time.

\section{Conclusion and future work}
We analyzed the effect of structural inductive biases (bags-of-words, sequences, graphs) on compositional generalization in referential games. We found that graph representations induce a stronger compositional prior measured by topographic similarity and out-of-domain generalization. Graph agents learn messages that lead to the highest accuracy in solving the task. Given the advancements in using graphs in natural language processing \citep{attention_nips}, a future direction could be to train graph speakers to generate sentences closer to natural language \citep{lowe*2020on}.

\bibliographystyle{plainnat}

\bibliography{cogsci}

\end{document}